%% file: m2942.tex
\documentclass[]{ecai}

\usepackage{latexsym}
\usepackage{amssymb}
\usepackage{amsmath}
\usepackage{amsthm}
\usepackage{booktabs}
\usepackage{enumitem}
\usepackage{graphicx}
\usepackage{color}

\usepackage{pgfplots}
\pgfplotsset{width=\textwidth,compat=1.9}

\usepackage{tabu}
\usepackage{multirow}
\usepackage{multicol}
\usepackage{float}
\usepackage{makecell}
\usepackage{tabularx}

\usepackage{amsfonts}
\usepackage{cancel}
\usepackage{bbding}

\usepackage{algorithm,algorithmic}

\usepackage{url}

\definecolor{red}{RGB}{254,76,97}
\definecolor{orange}{RGB}{243,156,17}
\definecolor{yellow}{RGB}{255,193,22}
\definecolor{green}{RGB}{82,196,26}
\definecolor{cyan}{RGB}{52,152,219}
\definecolor{blue}{RGB}{40,120,181}
\definecolor{purple}{RGB}{157,61,207}
\definecolor{black}{RGB}{14,29,105}
\definecolor{gray}{RGB}{191,191,191}

\newcommand{\BibTeX}{B\kern-.05em{\sc i\kern-.025em b}\kern-.08em\TeX}

\begin{document}


\begin{frontmatter}


\paperid{2942} 


\title{Router Upcycling: Leveraging Mixture-of-Routers in Mixture-of-Experts Upcycling}


\author[A]{\fnms{Junfeng}~\snm{Ran}\thanks{This work was conducted during an internship at Qiyuan Tech.}}
\author[B]{\fnms{Guangxiang}~\snm{Zhao}}
\author[A]{\fnms{Yuhan}~\snm{Wu}}
\author[A]{\fnms{Dawei}~\snm{Zhu}}
\author[A]{\fnms{Longyun}~\snm{Wu}}
\author[A]{\fnms{Yikai}~\snm{Zhao}}
\author[A]{\fnms{Tong}~\snm{Yang}}
\author[B]{\fnms{Lin}~\snm{Sun}\footnotemark[**]}
\author[B]{\fnms{Xiangzheng}~\snm{Zhang}\footnotemark[**]}
\author[A]{\fnms{Sujian}~\snm{Li}\thanks{Corresponding Authors. Email: \{sunlin1, zhangxiangzheng\}@360.cn, lisujian@pku.edu.cn}}

\address[A]{National Key Laboratory for Multimedia Information Processing, Peking University}
\address[B]{Qiyuan Tech}


\begin{abstract}
The Mixture-of-Experts (MoE) models have gained significant attention in deep learning due to their dynamic resource allocation and superior performance across diverse tasks.
However, efficiently training these models remains challenging.
The MoE upcycling technique has been proposed to reuse and improve existing model components, thereby minimizing training overhead.
Despite this, simple routers, such as linear routers, often struggle with complex routing tasks within MoE upcycling.
In response, we propose a novel routing technique called Router Upcycling to enhance the performance of MoE upcycling models.
Our approach initializes multiple routers from the attention heads of preceding attention layers during upcycling.
These routers collaboratively assign tokens to specialized experts in an attention-like manner.
Each token is processed into diverse queries and aligned with the experts' features (serving as keys).
Experimental results demonstrate that our method achieves state-of-the-art (SOTA) performance, outperforming other upcycling baselines.
\end{abstract}

\end{frontmatter}


\input{ecai/01-intro}
\input{ecai/02-preliminaries}
\input{ecai/03-method}
\input{ecai/04-experiments}
\input{ecai/05-analysis}
\input{ecai/06-ablation}
\input{ecai/07-related}
\input{ecai/08-conclusion}



\begin{ack}
We thank the anonymous reviewers for their helpful comments on this paper. 
This work was partially supported by National Natural Science Foundation of China projects 92470205 and 624B2005.
\end{ack}



\bibliography{custom}

\end{document}

%% file: ecai/01-intro.tex
\section{Introduction}
\label{sec: introduction}

\begin{figure}[ht]
\centering
\includegraphics[width=0.96\linewidth]{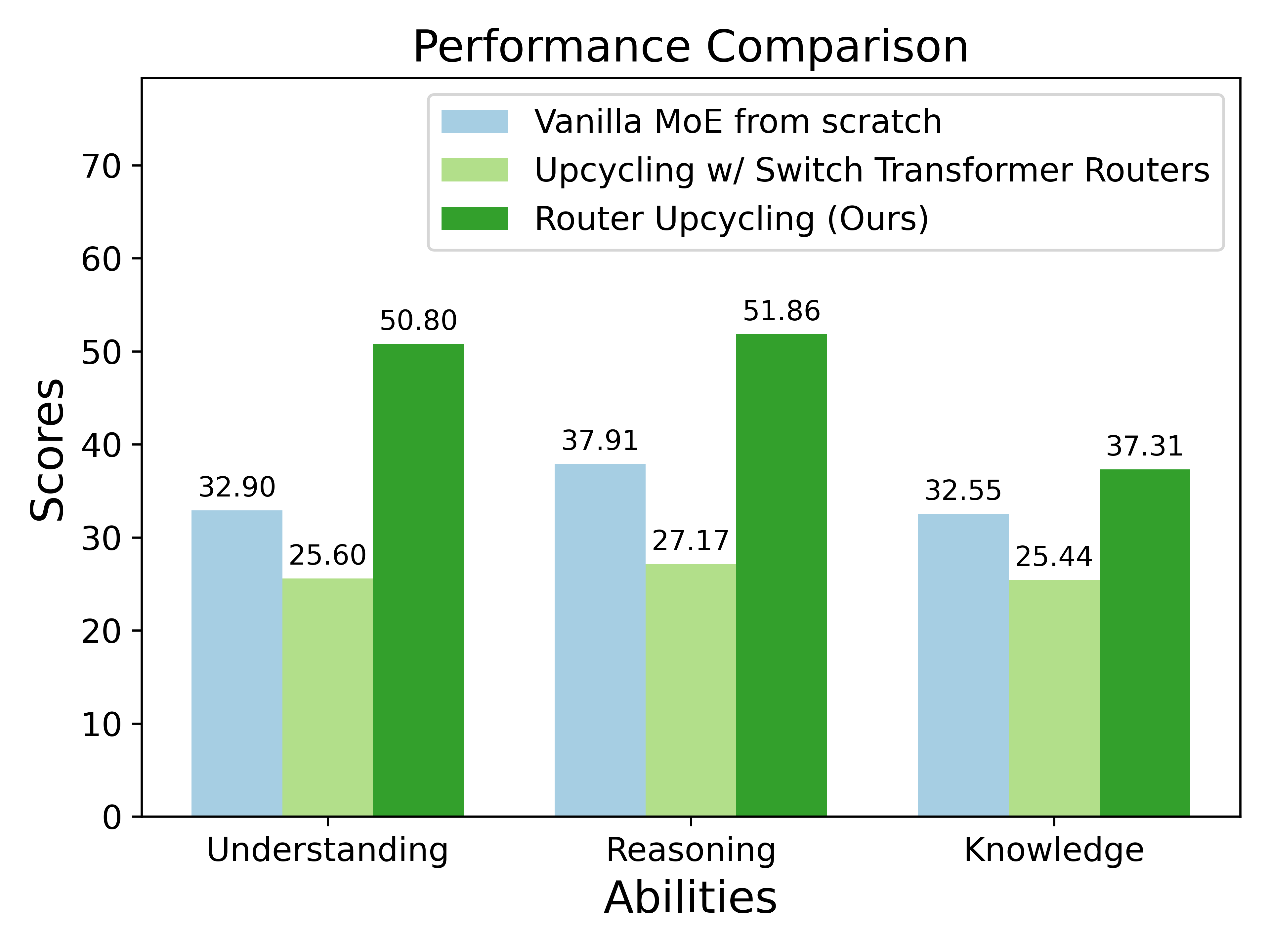}
\caption{Performance Comparison between vanilla MoE from scratch and various upcycling models with different routers on several benchmarks in Section~\ref{subsec: benchmarks}.}
\label{fig: bar_chart}
\end{figure}

\begin{figure*}[ht]
\centering
\includegraphics[width=0.96\linewidth]{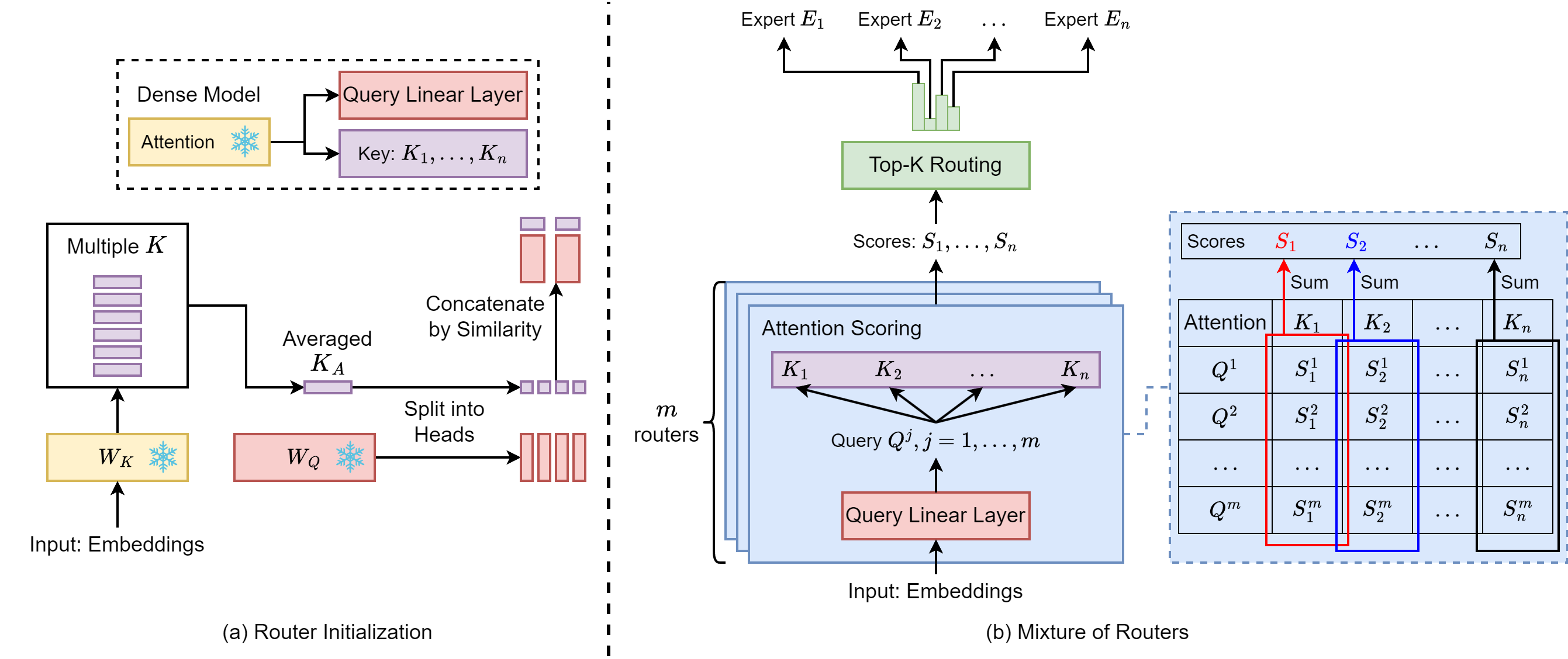}
\caption{Our proposed Router Upcycling framework. (a) Initializing routers with the Attention modules in a dense model, where \(W_Q\) and \(W_K\) are the query and key projections, and the "snowflakes" denote that parameters are frozen. (b) Computing routing scores $S$ between token queries $Q$ and expert keys $K$ using Attention scoring.}
\label{fig: method}
\end{figure*}

Mixture-of-Experts (MoE) models dynamically allocate computational resources, achieving strong performance across diverse tasks \cite{Jacobs1991AdaptiveMO, shazeerOutrageouslyLargeNeural2017, jiangMixtralExperts2024}.
By assigning input data to specialized expert networks via gating routers, MoE models enhance efficiency and accuracy.

The performance of MoE models depends heavily on routing strategies \cite{shazeerOutrageouslyLargeNeural2017}, as poor routers can lead to imbalanced expert training.
While vanilla MoE models use linear routers to assign tokens to top-\(k\) experts, these frameworks may fail to adapt to evolving expert structures, such as upcycling.
Upcycling \cite{komatsuzakiSparseUpcyclingTraining2023} is proposed as a popular approach that initializes experts from dense checkpoints, outperforming continued dense model training while reducing MoE training costs \cite{heUpcyclingLargeLanguage2024}.
Our preliminary experiments, as shown in Figure~\ref{fig: bar_chart}, present that leveraging inappropriate router structures in upcycling, such as upcycling with Switch Transformer \cite{fedusSwitchTransformersScaling2022} routers, negatively impact the performance of MoE upcycling models.
Therefore, exploring appropriate router structures is a crucial field for MoE upcycling.
However, building and initializing efficient routers remains to be explored in MoE upcycling.
In this paper, we examine the performance of previous router designs in MoE upcycling and propose a novel routing method for upcycling.
To our knowledge, we are the first to do router optimization in the Upcycling scenario.

Before we proceed, we need to determine why previous routers have failed.
Previous studies \cite{daiStableMoEStableRouting2022, liLocMoELowOverheadMoE2024, rollerHashLayersLarge2021} enhance the token-choice routing ability, many focusing on the ease of training instability that same or similar token representations are routed to different sets of experts during training \cite{daiStableMoEStableRouting2022}.
Nevertheless, most of them rely on fixed or simple structures like linear routers \cite{shazeerOutrageouslyLargeNeural2017}, and cannot inherit diverse token assignments, facing drawbacks such as the representation collapse issue \cite{chiRepresentationCollapseSparse2022}.
On the other hand, expert-choice routing \cite{zhouMixtureofExpertsExpertChoice2022} flips the routing paradigm by allowing experts to select tokens instead of tokens selecting experts, struggles in causal language modeling due to its reliance on future tokens.
More importantly, while experts are the same at the beginning of the MoE upcycling, different experts always choose the same tokens as input in expert-choice routing, where assigned tokens are not diverse enough for expert specialization.

In response, we propose Router Upcycling, a novel routing method (illustrated in Figure~\ref{fig: method}) that enhances MoE models by aligning a mixture of upcycled routers initialized from attention modules with the upcycled experts.
Specifically, each router leverages the query transformation from different attention heads in the preceding attention layer to process each token into diverse queries to highlight token representations from multiple perspectives.
Concurrently, during preprocessing, each expert's features are initialized as the average key computed by the attention heads in the preceding layer.
These expert features, serving as keys in the routers, are decoupled from token representations to ensure training stability.
In an attention-like manner, routing scores are calculated as the inner product of the diverse queries and the experts' keys, followed by a summation process to obtain the experts' scores, which are utilized for top-\(k\) routing afterward.
This attention-inspired mechanism captures inter-expert relationships, and tokens are routed based on multiple facets of their representations, which leads to more fine-grained token allocation.

Experimental results demonstrate that our method achieves state-of-the-art performance, consistently outperforming other upcycling baselines by over \(2\%\) on average.
Our routers add negligible computational overhead, making this enhancement a 'free lunch' for existing frameworks.
Furthermore, analytical experiments reveal that our routers produce more diverse token assignments, as evidenced by higher routing weight diversity for tokens.
Experts in our model also exhibit greater expert specialization, reflected in lower output similarity between expert pairs compared to vanilla upcycling.

The contribution of the paper is as follows.

\begin{itemize}[leftmargin=*]
    \item We are the first to do router optimization in the upcycling scenario, recognizing the necessity to upcycle routers and leverage attention modules to initialize them.
    
    \item We further expand the idea of upcycling by using Router Upcycling, where the MoE model is initialized without a random router as before, greatly enhancing the stability of MoE upcycling.
    
    \item Experimental evaluations indicate that our "free lunch" method achieves state-of-the-art performance, outperforming other upcycling baselines with a more than \(2\%\) improvement.
\end{itemize}

%% file: ecai/02-preliminaries.tex
\section{Preliminaries}
\label{sec: preliminaries}

\subsection{Mixture-of-Experts Layer}

A Mixture-of-Experts (MoE) layer typically consists of a set of \(n\) experts \(E_1, E_2, \dots, E_n\) and a router or gating network \(G\), which assigns tokens to the top-\(k\) experts \(E_s, s \in \mathbb{S}\) based on scores from a probability distribution calculated with the product of the input token and a gating weight matrix by an linear layer \cite{shazeerOutrageouslyLargeNeural2017}.
Let \(x \in \mathbb{R}^{d}\) be the input token representation, \(W\) be the linear layer that outputs an \(n\)-dimensional vector, and \(G(x)\) and \(E_i(x)\) be the outputs of the router and each expert \(E_i\), respectively.
The output \(y\) of the MoE layer, without normalization, can be written as follows:
\begin{equation}
    G(x) = \operatorname{Softmax} W(x).
\end{equation}
\begin{equation}
    \label{equ: moe}
    y = \sum_{s \in \mathbb{S}} G(x)_sE_s(x).
\end{equation}

\subsection{Upcycling}

The vanilla upcycling method \cite{komatsuzakiSparseUpcyclingTraining2023} initializes an MoE model from a dense model checkpoint.
It converts a dense model into an MoE model by duplicating the FFN weights multiple times and initializing a randomized router:
\begin{equation}
    \label{equ: upcycling}
    E_1 = E_2 = \dots = E_n = \operatorname{Dense} \operatorname{FFN}.
\end{equation}

%% file: ecai/03-method.tex
\section{Router Upcycling}
\label{sec: router upcycling}

\subsection{Overview}

The widely used vanilla linear router \cite{shazeerOutrageouslyLargeNeural2017} is too simple to handle diverse token assignments, leading to representation collapse \cite{chiRepresentationCollapseSparse2022}.
To address this, we introduce attention-like routers for better alignment between tokens and experts, as shown in Figure~\ref{fig: method}, where token embeddings act as queries and expert features act as keys.
Each token is transformed into multiple queries with different representations.
For example, one query may represent the syntax of a token, which is then matched with expert features using attention scoring.
This approach allows each query from a token to have a different semantic expression in a specific router subspace for more fine-grained and precise token assignments.
Considering multiple queries, our method constructs an equal number of routers, which work with a Mixture-of-Routers mechanism to collect the routing scores.

This section presents our novel Router Upcycling method for MoE models.
Following \cite{komatsuzakiSparseUpcyclingTraining2023}, our method initializes each expert as a copy of the original dense model's Feed-Forward Networks (FFN), as shown in Equation~\ref{equ: upcycling}, while keeping the dense model's other parts unchanged.
Our upcycled routers employ novel mechanisms:

\begin{enumerate}[leftmargin=*]
    \item \textbf{Multiple Routers Initialization from Attention Layers}: Initializing routers' query transformations and experts' keys from attention heads in the preceding attention layers, highlighting token representations from diverse perspectives.
    
    \item \textbf{Mixture-of-Routers Attention Scoring}: Using an attention-like mechanism to compute matching scores between token queries and expert keys, effectively aligning tokens with experts and incorporating scores from multiple collaborative routers to ensure diverse token routing.
\end{enumerate}

\subsection{Multiple Routers Initialization from Attention Layers}

As illustrated in Figure~\ref{fig: method}, this section describes how to initialize routers using attention heads from a dense model.
For simplicity, we set the number of routers \(m\) equal to the number of experts \(n\), with each expert matched to a single key.
This configuration is optimal in our experiments, and other settings and their corresponding results are discussed in Section~\ref{sec: number_ablation}.

Our method performs an additional pre-training process during which the dense model is frozen to initialize the routers and experts.
During this phase, we extract query transformation matrices \(W_Q\) and average attention key representations \(K_A\) from the dense model’s \(h\) split attention heads.
The detailed process will be elaborated in Section~\ref{subsec: setup}.

Since the dimensionality of individual attention heads may be insufficient for effective routing expressivity, we enhance their representation power by concatenating pairs of query transformations and key representations. 
As empirically validated in Section~\ref{sec: number_ablation} and Table~\ref{tab: number}, the single-head dimensions proved to be insufficient, and concatenation effectively doubled their routing expressivity.
We utilize a greedy approach for concatenation as it has been shown to outperform alternative strategies.
For instance, preliminary tests with a random concatenation strategy resulted in a minimal yet consistent performance degradation of approximately 0.5\%.
Specifically, we perform this concatenation \(log_2\ h/m\) times, using a greedy search algorithm that selects pairs with the highest cosine similarity.
This process yields concatenated query transformations \(W_Q^C\) and key representations \(K_A^C\), having larger hidden dimensions and improved expressiveness.

Next, we use \(W_Q^C\) and \(K_A^C\) (highlighted in red and purple in Figure~\ref{fig: method}(a)) to initialize the token query transformations \(W\) and expert keys \(K\) (highlighted in red and purple in Figure~\ref{fig: method}(b)) in our proposed mixture of \(m\) routers and \(n\) experts:
\begin{equation}
    \text{Expert}_i: K_i = (K_A^C)_i, \quad i = 1, \dots, n,
\end{equation}
\begin{equation}
    \text{Router}^j: W^j = (W_Q^C)^j, \quad j = 1, \dots, m,
\end{equation}

The routers transform each token embedding \(x\) by projecting it into \(m\) low-dimensional subspaces using linear transformations:
\begin{equation}
    Q^j = W^jx, \quad j = 1, \dots, m,
\end{equation}
where \(Q^j \in \mathbb{R}^{d'}\) is the \(j\)-th query for token embedding \(x\) in the \(j\)-th router, and \(W^j \in \mathbb{R}^{d' \times d}\) is the projection matrix.
Unlike tokens, each expert \(E_i\) preserves its unique key embedding \(K_i \in \mathbb{R}^{d'}\), independent of the token representation, to maintain its stable feature.

\subsection{Mixture-of-Routers Attention Scoring}

With the token queries and expert keys built, our method routes tokens in an attention-like manner.
Unlike the traditional attention mechanism \cite{vaswani2023attention}, these queries are from the same token instead of a sequence of tokens, and the keys, held by experts, are independent of tokens.
The attention scoring obtains attention-mapping scores by multiplying each token query \(Q^j\) in \(m\) routers by each expert key \(K_i\) for each \(Q\)-\(K\) pair, as shown on the right side of Figure~\ref{fig: method}:
\begin{equation}
    S_i^j = \frac{Q^j{}^\top K_i}{\sqrt{d'}}, i = 1, \dots, n, j = 1, \dots, m,
\end{equation}
where \(S_i^j\) represents the matching score between the \(j\)-th query of the token and the \(i\)-th key of expert \(E_i\).
To maintain the diverse amplitude of each low-dimensional \(Q\)-\(K\) subspace, our method does not use any normalization technique such as the cosine router \cite{chiRepresentationCollapseSparse2022}.

To incorporate attention-mapping scores from collaborative routers, we sum over the query dimension for each token:
\begin{equation}
    S_i = \sum_{j} S_i^j, \quad i = 1, \dots, n, j = 1, \dots, m.
\end{equation}
Finally, to obtain the top-\(k\) routing weights \(R\) for expert \(E_i\), these scores are sent to a top-\(k\) router:
\begin{equation}
    R = \operatorname{Softmax}(S).
\end{equation}
By selecting the top-\(k\) experts with the highest routing weights \(R\), we assign tokens to the most appropriate experts and use Equation~\ref{equ: moe} to get the output of the experts.

%% file: ecai/04-experiments.tex
\section{Experiments}
\label{sec: experiment}

\begin{table*}[ht]
\caption{Performance comparison of different models on benchmark datasets, evaluated with zero-shot schema on every benchmark.}
\centering
\small
\begin{tabular}{lccccccc}
\toprule
\multirow{2}{*}{\textbf{Model}} & Vanilla MoE & Vanilla & Switch & LocMoE & Upcycling & \textbf{Router} \\
 & from scratch & Upcycling & Upcycling & Upcycling & w/ MLP routers & \textbf{Upcycling} \\
\midrule
OBQA & 31.4 & 40.8 & 27.6 & 38.2 & 37.6 & \textbf{42.6} \\
OBQA-fact & 34.4 & 58.0 & 23.6 & 51.4 & 55.0 & \textbf{59.0} \\
\midrule
ARC-C & 27.12 & 39.66 & 14.58 & 29.15 & 34.92 & \textbf{42.37} \\
ARC-E & 30.51 & 52.73 & 20.63 & 45.68 & 51.15 & \textbf{58.02} \\
Hellaswag & 41.45 & 49.36 & 25.24 & 48.83 & 49.70 & \textbf{50.12} \\
Winogrande & 52.57 & 56.67 & 48.22 & 56.12 & \textbf{58.88} & 56.91 \\
\midrule
BoolQ & 48.13 & 49.88 & 47.09 & 54.19 & 41.44 & \textbf{55.84} \\
COPA & 60 & 62 & 49 & 61 & 63 & \textbf{64} \\
NQ & 4.32 & 4.85 & 0.42 & 4.04 & 3.96 & \textbf{5.15} \\
TriviaQA & 17.74 & 23.82 & 5.24 & 22.16 & 24.17 & \textbf{24.25} \\
\midrule
Understanding Average & 32.90 & 49.40 & 25.60 & 44.80 & 46.30 & \textbf{50.80} \\
Reasoning Average & 37.91 & 49.61 & 27.17 & 44.95 & 48.67 & \textbf{51.86} \\
Knowledge Average & 32.55 & 35.14 & 25.44 & 35.35 & 33.14 & \textbf{37.31} \\
\midrule
Average & 34.76 & 43.78 & 26.17 & 41.08 & 41.98 & \textbf{45.83} \\
\bottomrule
\end{tabular}
\label{tab: mainresults}
\end{table*}

\begin{figure}[ht]
\centering
\includegraphics[width=0.96\linewidth]{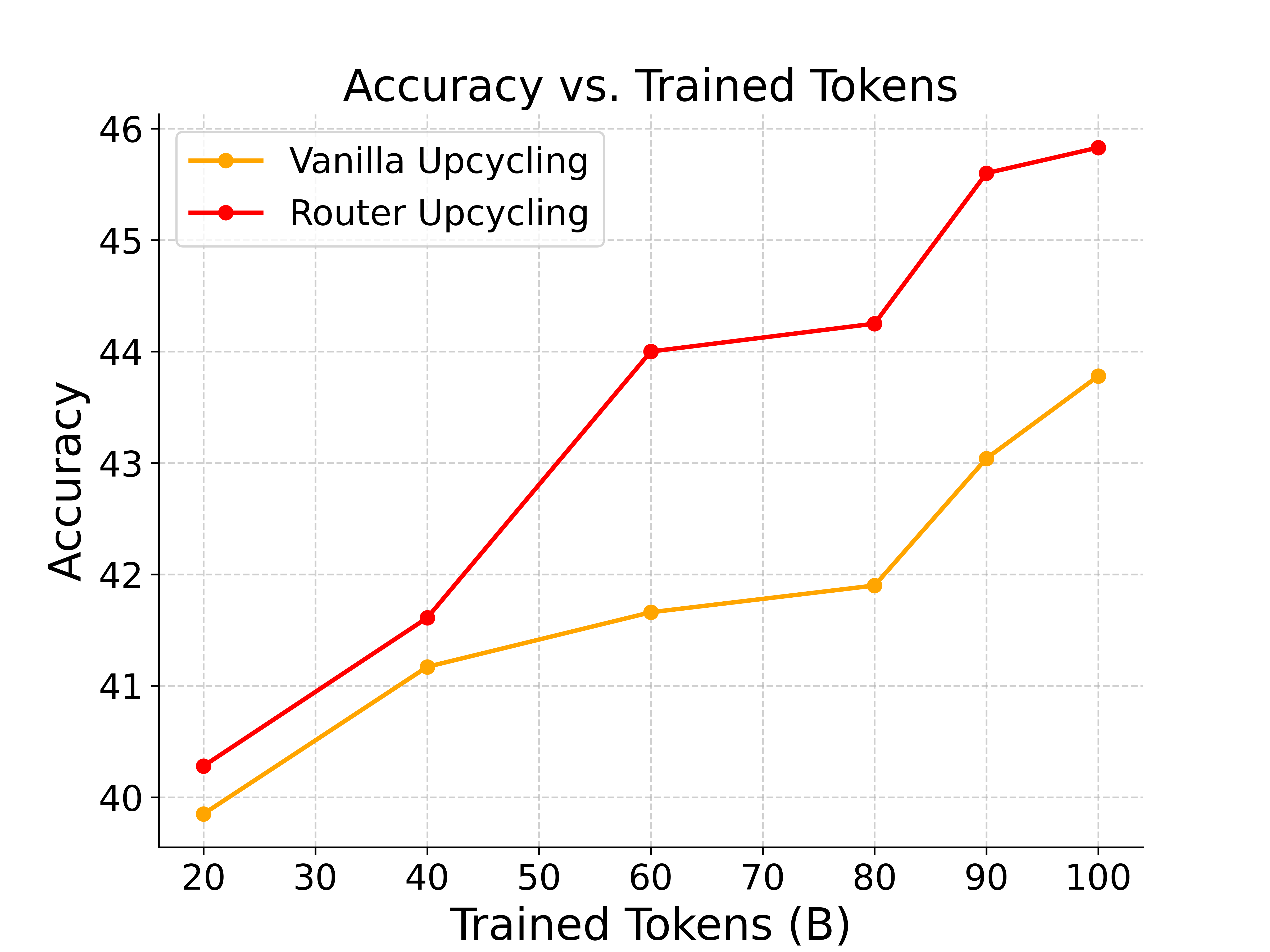}
\caption{Average accuracy comparison during training.}
\label{fig: accuracy}
\end{figure}

\begin{figure}[ht]
\centering
\includegraphics[width=0.96\linewidth]{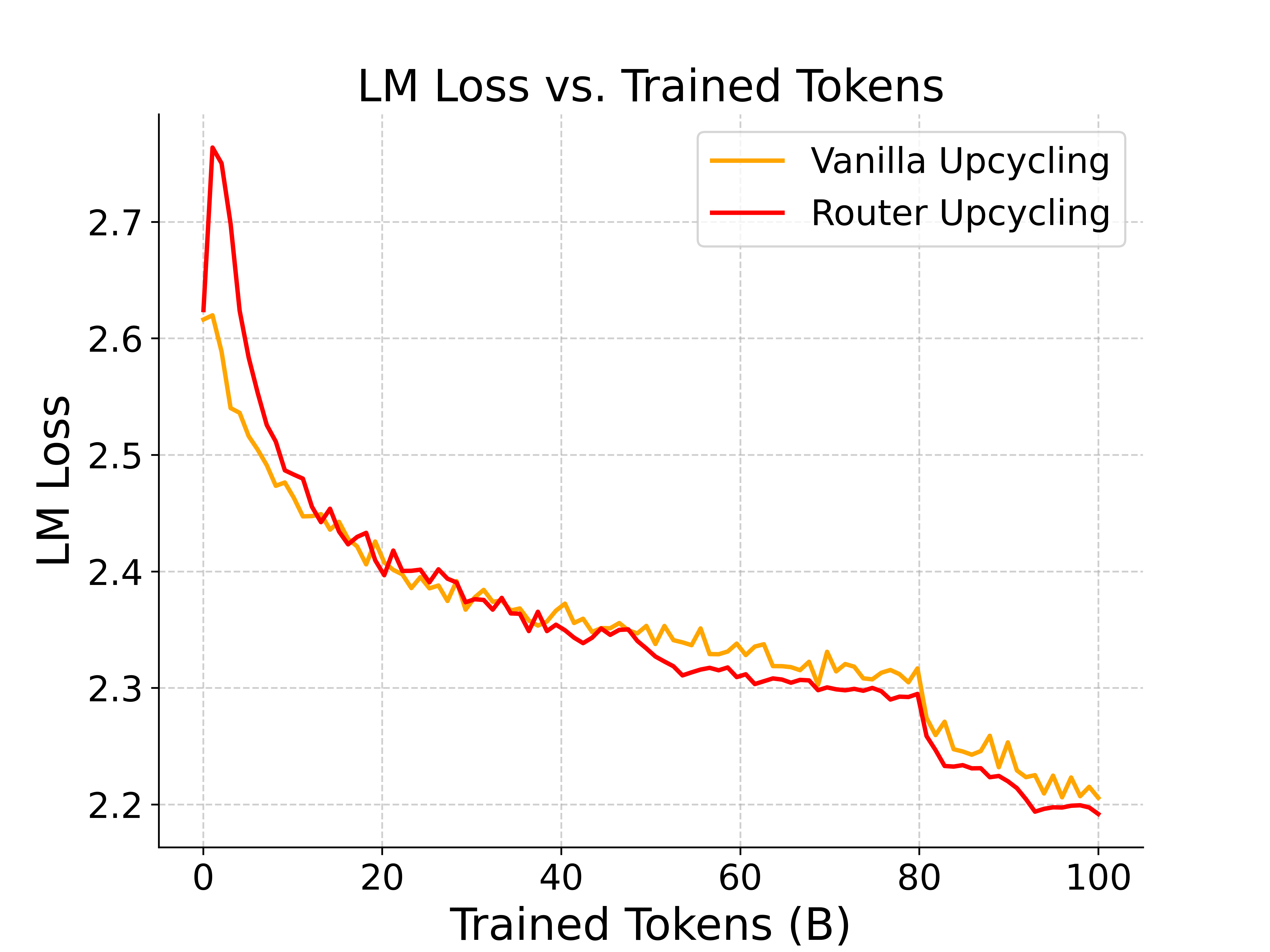}
\caption{LM loss comparison during training.}
\label{fig: loss}
\end{figure}

\subsection{Experimental Setup}
\label{subsec: setup}

We conducted our main experiments on an upcycled Qwen 8x0.5B model \cite{qwen} with approximately 2.1B total parameters and around 0.8B activated parameters.
Our architecture maintains the Qwen models' structure and aligns with Mixtral's established practice (8 experts/layer). 
These hyperparameters were set using standard practices from prior work and were further confirmed through preliminary experiments. 
We provide additional validation of these choices through the ablation studies presented in Section~\ref{sec: number_ablation} and Table~\ref{tab: number}.
Our experiments were conducted under the Megatron \cite{shoeybi2020megatronlmtrainingmultibillionparameter} framework on 64 NVIDIA H-100 GPUs.
Since the model size is relatively small, each GPU has a copy of the whole model to ensure computational balance, and the micro-batch size is set to 4.
Models are trained on 100B tokens, sampled from a large-scale multilingual corpus \cite{gao2020pile800gbdatasetdiverse, weber2024redpajamaopendatasettraining} designed for continued pretraining.
In section~\ref{sec: scaling}, we also conducted scaling-up experiments with a Qwen 8x1.8B model and a Qwen 16x0.5B model.

Taking the 8x0.5B model as an example, before training, we upcycled the dense model by duplicating the FFN 8 times to form \(n=8\) experts and apply a top-2 selection, which is a classic setting \cite{jiangMixtralExperts2024} and initializing \(m=8\) of our proposed routers in each layer to construct an 8x0.5B MoE model.
Other parts of the dense model remained unchanged.
The average key representations in the router were obtained by freezing the dense model and calculating the average key vectors over 10 iterations, with a batch size of 1024 and a sequence length of 4096.
The original dense model employs \(h=16\) attention heads, each with a dimension of 64, resulting in a total feature dimension of \(d=1024\).
We concatenate every 2 token query transformations and every 2 expert keys using a greedy search algorithm that selects pairs with the highest cosine similarity, to form \(m=8\) routers to align with the expert number \(n=8\), and the intermediate dimension of each router is \(d'=128\).
The additional router parameters are approximately \(8 \times 1024 \times 128 = 1\text{M}\) for each layer, which is negligible compared to the 2.1B total MoE parameters.

During training, we employ the Adam optimizer \cite{kingma2017adammethodstochasticoptimization} with hyper-parameters set to \(\beta_1=0.9\), \(\beta_2=0.95\), \(\epsilon=10^{-8}\), gradient clipping norm \(=1.0\) and weight decay \(=0.1\).
We do not use dropout due to the abundant training corpus.
The learning rate is scheduled using a warmup-and-step-decay strategy \cite{daiDeepSeekMoEUltimateExpert2024}.
In the first \(1\%\) of warm-up steps, the learning rate increases from 0 to the maximum value, which is set to \(5 \times 10^{-4}\).
The learning rate stays at this constant value until the last \(20\%\) of training steps, where it is multiplied by 0.316 (approximately \(1/\sqrt{10}\)) at \(80\%\) and \(90\%\) of the training steps.
Each training batch contains 4M tokens, with the batch size and sequence length set to 1024 and 4096, respectively.
The total number of training steps is 25600 to match 100B training tokens.

For MoE settings, we do not drop any tokens during training except for the Switch Transformer \cite{fedusSwitchTransformersScaling2022}.
Our model leverages an auxiliary loss \cite{lewisBASELayersSimplifying2021} of 0.02 and a router z-loss \cite{zophSTMoEDesigningStable2022} of 0.001 to improve router stability.

\subsection{Evaluation Benchmarks}
\label{subsec: benchmarks}

We conduct experiments on several benchmark datasets commonly used in evaluating MoE models, grouped by the abilities needed:

\begin{itemize}[leftmargin=*]
    \item Understanding: OpenbookQA (short as OBQA), OpenbookQA-fact \cite{mihaylov2018suitarmorconductelectricity}.
    
    \item Reasoning: ARC-C \cite{clark2018thinksolvedquestionanswering} and ARC-E, Hellaswag \cite{zellers2019hellaswagmachinereallyfinish}, Winogrande \cite{sakaguchi2019winograndeadversarialwinogradschema}.

    \item Knowledge: BoolQ \cite{clark2019boolqexploringsurprisingdifficulty}, COPA \cite{Gordon2011ChoiceOP}, TriviaQA \cite{joshi2017triviaqalargescaledistantly}, NQ \cite{kwiatkowski-etal-2019-natural}
\end{itemize}

We evaluate models in a zero-shot manner to showcase their generalization abilities without further instructions.

\subsection{Baselines}

We compare Router Upcycling with five other MoE model variants:

\begin{itemize}[leftmargin=*]
    \item \textbf{Vanilla MoE from scratch}: A vanilla MoE model \cite{shazeerOutrageouslyLargeNeural2017} using traditional softmax gating.
    A normal initialization (mean \(=0.0\), std \(=0.02\)) is applied to initialize all parameters.
    
    \item \textbf{Vanilla Upcycling}: Based on the vanilla MoE model, experts' parameters are upcycled \cite{komatsuzakiSparseUpcyclingTraining2023} as copies of the original dense model's FFN, and other parameters are also converted from the dense model, except routers.
    
    \item \textbf{Switch Upcycling}: Based on the Vanilla Upcycling model, the router is changed to the one proposed in Switch Transformer \cite{fedusSwitchTransformersScaling2022}.
    
    \item \textbf{LocMoE Upcycling}: Based on the Vanilla Upcycling model, the router is changed to the one proposed in LocMoE \cite{liLocMoELowOverheadMoE2024}.
    
    \item \textbf{Upcycling w/ MLP routers}: Based on the Vanilla Upcycling model, a two-layer MLP with a GELU \cite{hendrycks2023gaussianerrorlinearunits} activation function is used as the router, with the following dimension transformations: Layer 1: 1024 \(\rightarrow\) 1024; Layer 2: 1024 \(\rightarrow\) 8. 
    Its router parameters are approximately the same as the proposed method.
\end{itemize}

%% file: ecai/05-analysis.tex
\section{Results and Analysis}
\label{sec: analysis}

\subsection{Main Results}

Table~\ref{tab: mainresults} presents the performance comparison between our proposed method, Router Upcycling, and various upcycling baselines.
Router Upcycling achieves the highest average performance across all benchmarks, with an average score of 45.83, representing a significant improvement of 2.05 points over the Vanilla Upcycling method.
This consistent outperformance highlights our proposed routing mechanism's robustness and generalization capabilities.

Our method also accelerates the evolution of the MoE upcycling more effectively than Vanilla Upcycling, as illustrated in Figure~\ref{fig: accuracy} and Figure~\ref{fig: loss}.
Figure~\ref{fig: accuracy} shows that Router Upcycling consistently outperforms Vanilla Upcycling throughout the entire training process, with particularly notable gains in the middle training stages. 
Figure~\ref{fig: loss} further highlights the advantages of Router Upcycling in terms of loss dynamics during training.
While downstream task performance can often benefit from rewarming the model with a large learning rate, despite the associated temporary rise in loss during the warmup stage \cite{gupta2023continualpretraininglargelanguage}, the loss curve for Vanilla Upcycling remains flat, indicating insufficient warmup.
In contrast, Router Upcycling effectively leverages the warmup phase and gains faster convergence (stable loss at 50B tokens vs. baseline fluctuations) in the first step-decay learning stage.
Consequently, after the 50B training schedule, our Router Upcycling model achieves lower loss and higher downstream task performance.

\subsection{Scaling Up the Model}
\label{sec: scaling}

\begin{table*}[ht]
\caption{Performance comparison of models scaling up with a larger backbone model or more experts.}
\centering
\small
\begin{tabular}{lcccccc}
\toprule
\multirow{2}{*}{\textbf{Model}} & 8x0.5B Vanilla & 8x0.5B Router & 8x1.8B Vanilla & 8x1.8 Router & 16x0.5B Vanilla & 16x0.5B Router \\
 & Upcycling & Upcycling & Upcycling & Upcycling & Upcycling & Upcycling \\
\midrule
OBQA & 40.8 & 42.6 & 42.2 & 47.4 & 34.8 & 37.0 \\
OBQA-fact & 58.0 & 59.0 & 59.6 & 63.8 & 50.8 & 54.2 \\
\midrule
ARC-C & 39.66 & 42.37 & 42.71 & 44.07 & 36.61 & 40.00 \\
ARC-E & 52.73 & 58.02 & 58.73 & 56.97 & 55.56 & 56.08 \\
Hellaswag & 49.36 & 50.12 & 54.25 & 57.95 & 49.27 & 50.00 \\
Winogrande & 56.67 & 56.91 & 59.12 & 61.8 & 56.67 & 57.62 \\
\midrule
BoolQ & 49.88 & 55.84 & 47.92 & 43.94 & 38.84 & 41.01 \\
COPA & 62 & 64 & 68 & 67 & 58 & 61 \\
NQ & 4.85 & 5.15 & 5.01 & 6.59 & 3.46 & 3.66 \\
TriviaQA & 23.82 & 24.25 & 28.29 & 32.92 & 23.82 & 23.93 \\
\midrule
Average & 43.78 & 45.83 & 46.58 & 48.24 & 40.78 & 42.45 \\
\bottomrule
\end{tabular}
\label{tab: scaling}
\end{table*}

Table~\ref{tab: scaling} compares model performance when scaling up the backbone model size and number of experts.
Key observations include: (1) Router Upcycling remains effective at larger scales, and (2) the performance gains from scaling up are marginal or negative for upcycling, particularly when increasing the number of experts.

Across all configurations, Router Upcycling consistently outperforms Vanilla Upcycling, demonstrating its robustness and adaptability even as the model scales.
For instance, in the 8x0.5B configuration, Router Upcycling achieves an average score of 45.83, surpassing Vanilla Upcycling’s 43.78 by 2.05 points.
Similarly, in the 8x1.8B configuration, Router Upcycling achieves an average score of 48.24, outperforming Vanilla Upcycling’s 46.58 by 1.66 points.

While scaling up the backbone model size improves performance, the gains are relatively modest for both upcycling methods.
For instance, the average score for Vanilla Upcycling increases from 43.78 in the 8x0.5B configuration to 46.58 in the 8x1.8B configuration, a gain of only 2.80 points.
Similarly, Router Upcycling improves from 45.83 to 48.24, a gain of 2.41 points.

More notably, increasing the number of experts from 8 to 16 results in marginal or even negative performance gains.
For example, the average score for Vanilla Upcycling drops from 43.78 in the 8x0.5B configuration to 40.78 in the 16x0.5B configuration.
Similarly, Router Upcycling decreases from 45.83 to 42.45 under the same configurations. 
This indicates that scaling up the backbone model or the number of experts does not necessarily lead to better performance in upcycling frameworks.
The drops are more likely to stem from the 16-expert models' limited head dimensions that constrain routing capacity.
Our future work will explore initializing heads with diverse corpora multiple times to expand dimensionality, where head dimensionality would not be constrained.

\subsection{Routing Diversity}
\label{subsec: diversity}

\begin{figure*}[ht]
    \centering
    \begin{minipage}{1\linewidth}
        \centering
        \includegraphics[width=0.8\linewidth]{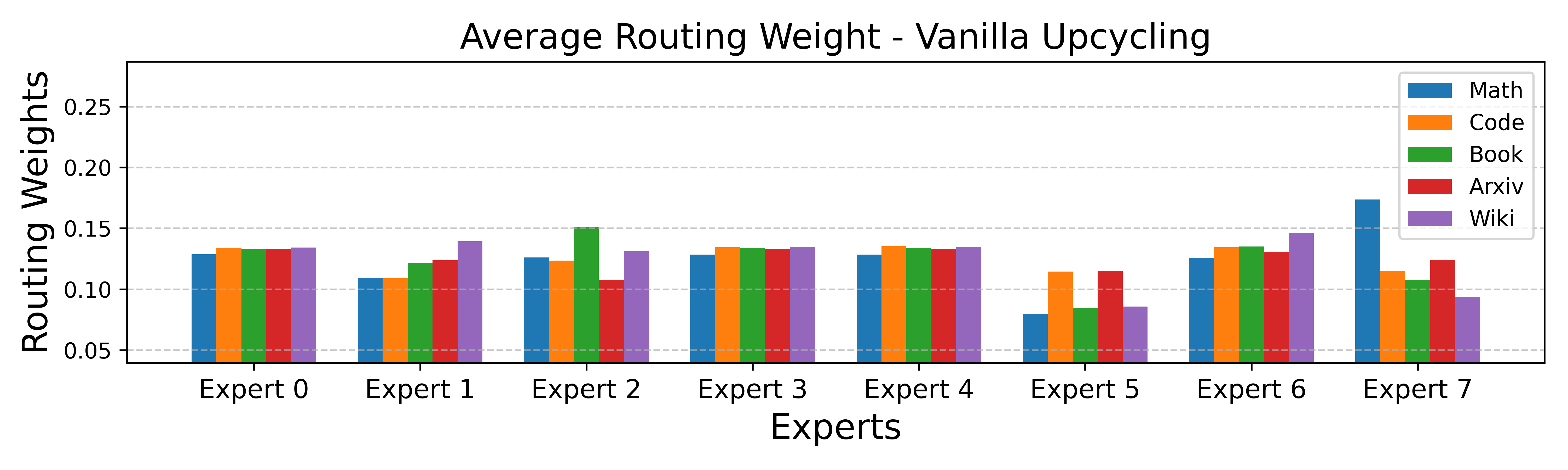}
    \end{minipage}
    
    \begin{minipage}{1\linewidth}
        \centering
        \includegraphics[width=0.8\linewidth]{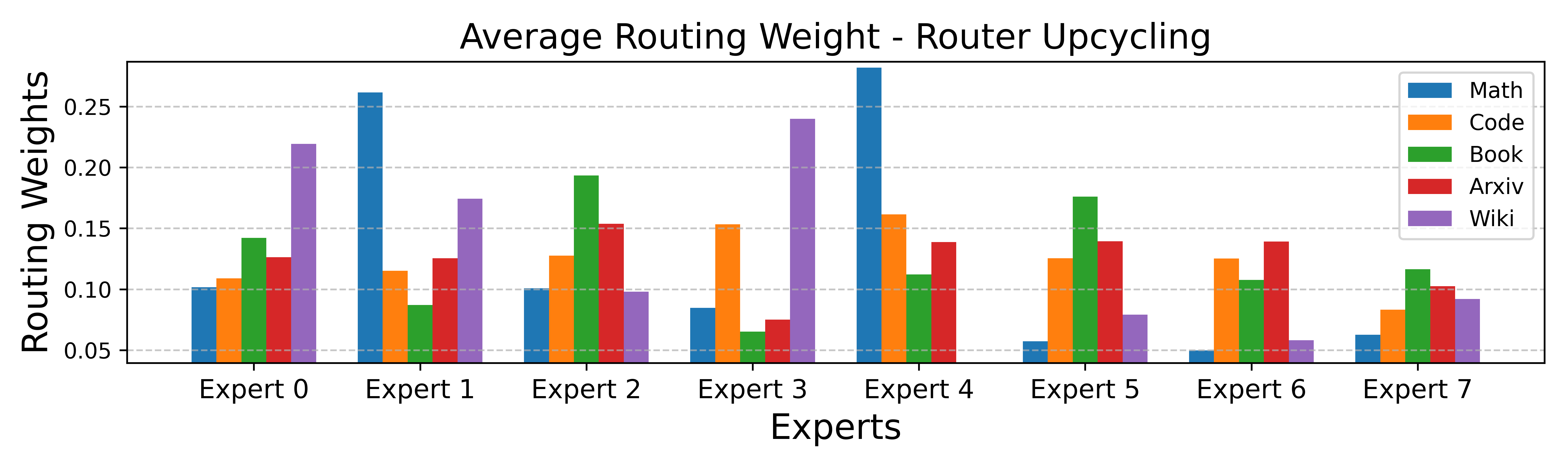}
    \end{minipage}
    \caption{Average routing weights of Vanilla Upcycling and Router Upcycling in the last layer (layer 23) of the 8x0.5B MoE models.}
    \label{fig: routing_weight}
\end{figure*}


To assess the diversity of token assignments in Router Upcycling compared to Vanilla Upcycling, we analyze the average routing weights of experts across five domains: Math, Code, Book, Arxiv, and Wiki \cite{gao2020pile800gbdatasetdiverse}.
Routing diversity is measured by routing weights across experts, where a more uneven distribution indicates greater diversity in token assignments.
Figure~\ref{fig: routing_weight} illustrates the routing weight distribution in the last layer (layer 23).

Vanilla Upcycling exhibits low routing diversity, with weights distributed relatively evenly across experts for all domains.
For example, in the Code domain, weights range from 0.1090 to 0.1353, which is quite narrow, and experts such as 0, 3, and 4 show flat preferences across all five domains.
This uniform distribution suggests that Vanilla Upcycling struggles to achieve diverse token assignments.
In contrast, Router Upcycling demonstrates significantly higher routing diversity, with uneven weight distributions across experts.
For instance, in the Math domain, Expert 4 has a peak weight of 0.2818, while Expert 6 receives a much lower weight of 0.0498.
Similarly, in the Wiki domain, Expert 3 dominates with a weight of 0.2398, while Expert 4 has a minimal weight of 0.0394.
These pronounced differences in routing weights across domains highlight Router Upcycling to assign tokens more diversely, enabling experts to specialize in specific tasks or domains.

\subsection{Expert Specialization}
\label{subsec: specialization}

\begin{figure}[ht]
\centering
\includegraphics[width=0.96\linewidth]{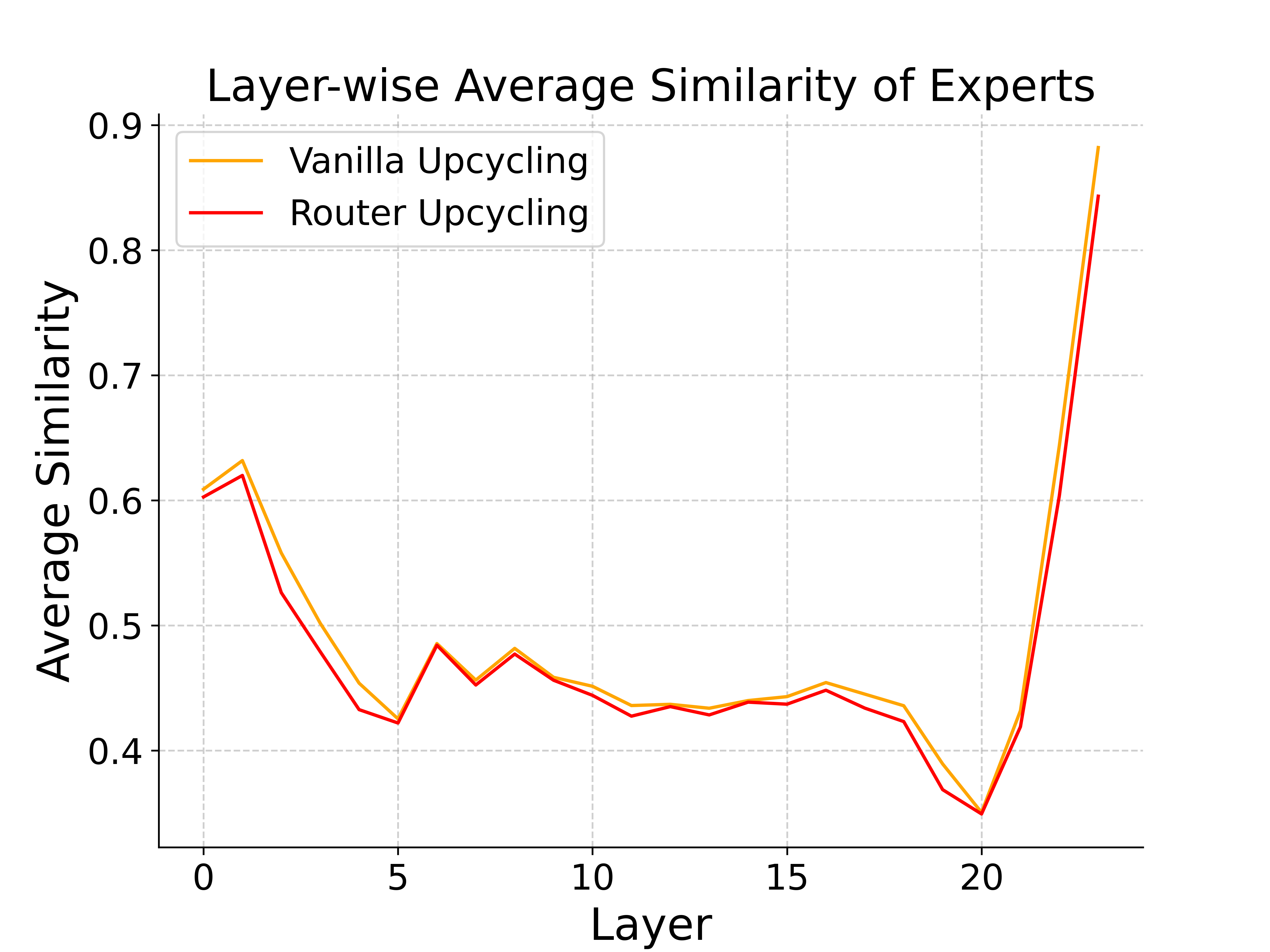}
\caption{Layer-wise average output cosine similarity of 8x0.5B MoE models' experts.}
\label{fig: similarity}
\end{figure}

We measure the layer-wise average output cosine similarity between expert pairs to evaluate expert specialization.
Lower cosine similarity indicates greater specialization, reflecting more distinct outputs from experts.
Figure~\ref{fig: similarity} shows that Router Upcycling consistently achieves lower cosine similarity across layers, indicating enhanced expert specialization.
For instance, in deeper layers such as Layer 22 and 23, Router Upcycling reduces cosine similarity by \(-4.01\%\) and \(-3.89\%\), respectively.
Similarly, in earlier layers like Layer 2 and 3, reductions of \(-3.17\%\) and \(-2.27\%\) are observed.
These improvements are particularly notable in the earlier and deeper layers, where specialized experts are essential in processing and generating task-specific tokens.
The results demonstrate that Router Upcycling fosters greater expert specialization, enabling experts to produce more distinct and specialized outputs across all layers.

%% file: ecai/06-ablation.tex
\section{Ablation Study}

\subsection{Number of Routers, Keys, and Experts}
\label{sec: number_ablation}

\begin{table*}[ht]
\caption{Ablation study on the number of routers, keys, and experts.
\(m\) is the router number, \(keys\) is the number of expert keys and \(dim\) is their dimension, and \(n\) is the expert number.
Note that we have conducted all possible ablation studies on the numbers.}
\centering
\small
\begin{tabular}{lcccccc}
\toprule
\textbf{Model} & Vanilla Upcycling & \(m\)=2, \(n\)=8 & \(m\)=4, \(n\)=8 & \textbf{\(m\)=\(n\)=8} & \(m\)=16, \(n\)=8 & \(m\)=32, \(n\)=8 \\
\midrule
\(keys\) & - & 8 & 8 & 8 & 16 & 32 \\
\(dim\) & - & 512 & 256 & 128 & 64 & 32 \\
\midrule
OBQA & 40.8 & 38.0 & 29.8 & \textbf{42.6} & 26.8 & 42.2 \\
OBQA-fact & 58.0 & 57.8 & 40.8 & \textbf{59.0} & 27.6 & 52.6 \\
\midrule
ARC-C & 39.66 & 32.88 & 34.58 & \textbf{42.37} & 23.73 & 41.02 \\
ARC-E & 52.73 & 45.21 & 40.92 & \textbf{58.02} & 23.81 & 51.15 \\
Hellaswag & 49.36 & 47.24 & 39.56 & \textbf{50.12} & 30.67 & 49.26 \\
Winogrande & 56.67 & 57.62 & 55.56 & 56.91 & 52.49 & \textbf{58.48} \\
\midrule
BoolQ & 49.88 & 51.65 & 47.22 & \textbf{55.84} & 44.43 & 53.49 \\
COPA & 62 & \textbf{68} & 63 & 64 & 59 & 64 \\
NQ & 4.85 & 3.32 & 2.52 & \textbf{5.15} & 1.05 & 3.96 \\
TriviaQA & 23.82 & 18.78 & 13.45 & \textbf{24.25} & 9.39 & 22.8 \\
\midrule
\textbf{Average} & 43.78 & 42.05 & 36.74 & \textbf{45.83} & 29.90 & 43.90 \\
\bottomrule
\end{tabular}
\label{tab: number}
\end{table*}

In this section, we investigate the impact of varying the number of routers, expert keys, and experts on model performance in Table~\ref{tab: number}.
The router number \(m\) is a critical hyperparameter as it determines how attention heads are paired and concatenated in the base model.
Two rules are applied to form proper models when the number of routers \(m\) and keys \(keys\) changes compared to a fixed number of experts \(n\):

\begin{enumerate}[leftmargin=*]
    \item If the router number \(m < n\), attention heads are duplicated \(n/m\) times before concatenation to match the hidden size in the routers, and each expert holds one key, so \(keys = n\).

    \item If the router number \(m > n\), without any duplication, \(m = keys\), and each expert preserves more than one key as their feature representation, which would be selected based on the top-k score of any of its keys.
Additionally, we create a new model variant ``\(m\)=32, \(n\)=8'' by splitting each attention head in the original dense model into two to explore further possibilities.
\end{enumerate}

The results indicate no particular trend when the router number \(m\) changes, but the model performs best when \(m\)=\(n\)=8.
Intuitively, aligning the number of routers \(m\) with the number of experts \(n\) appears to benefit the routing process, which is empirically validated.
When \(m=16\), the performance is the worst, possibly due to the limited expression power of a single attention head with a dimension of only 64 for the routing task.
However, the model performance improves when \(m=32\) despite the smaller attention head dimension.

\subsection{Router Mixture Methods}
\label{sec: mixture}

\begin{table}[ht]
\caption{Performance comparison between two models utilizing max pooling and summation as mixture methods.}
\centering
\small
\begin{tabular}{lcc}
\toprule
\textbf{Method} & Max Pooling & \textbf{Summation} \\
\midrule
OBQA & \textbf{44.6} & 42.6 \\
OBQA-fact & 56.4 & \textbf{59.0} \\
\midrule
ARC-C & 37.63 & \textbf{42.37} \\
ARC-E & 54.14 & \textbf{58.02} \\
Hellaswag & 49.55 & \textbf{50.12} \\
Winogrande & \textbf{58.17} & 56.91 \\
\midrule
BoolQ & \textbf{56.54} & 55.84 \\
COPA & 63 & \textbf{64} \\
NQ & 3.38 & \textbf{5.15} \\
TriviaQA & 22.62 & \textbf{24.25} \\
\midrule
\textbf{Average} & 44.60 & \textbf{45.83} \\
\bottomrule
\end{tabular}
\label{tab: mixture}
\end{table}

In this section, we explore alternative mixture methods for collaborative routers, comparing summation with max pooling to determine the optimal method for output aggregation.
Max pooling involves using the top-\(k\) router score of all Query-Key pairs in all routers to route the token.
As shown in Table~\ref{tab: mixture}, the summation method generally performs better than max pooling.

%% file: ecai/07-related.tex
\section{Related Work}
\label{sec: relatedwork}

Improving routers has been an important and interesting field for many years in the MoE study.
Prior works introduce noise and normalization to enhance the robustness of routers.
Shazeer et al.~\cite{shazeerOutrageouslyLargeNeural2017} proposed improvements by introducing noise for load balance and retaining the top-k experts.
Switch Transformers~\cite{fedusSwitchTransformersScaling2022} address overfitting in fine-tuning tasks with limited examples by simplifying the routing mechanism using a top-1 gating strategy, reducing computational overhead and communication costs.

Other works focus on adjusting routing mechanisms.
StableMoE \cite{daiStableMoEStableRouting2022} proposes a two-stage routing strategy with a distilled router for stable decisions.
Zuo et al.~\cite{zuoTamingSparselyActivated2022} introduce stochastic experts to bypass the router, promoting consistency through regularization.
LocMoE \cite{liLocMoELowOverheadMoE2024} introduces a GrAP layer that divides the hidden state of tokens, computes gating values without learnable parameters, and adds a locality loss to ensure tokens are preferentially routed to local experts.
Several studies have also explored dynamic routing mechanisms \cite{huangHarderTasksNeed2024, zengAdaMoETokenAdaptiveRouting2024}, allowing tokens to select a varying number of experts based on input difficulty, enhancing computational efficiency and model performance.

However, these routing mechanisms often encourage token clustering around expert centroids, leading to representation collapse \cite{chiRepresentationCollapseSparse2022}.
To tackle this, Chi et al.~\cite{chiRepresentationCollapseSparse2022} leverage dimension reduction using linear projection to isolate interactions on a low-dimensional hypersphere.
Other novel routing methods include expert choice routing \cite{zhouMixtureofExpertsExpertChoice2022}, where experts select tokens, and hashing-based routing \cite{rollerHashLayersLarge2021}, replacing traditional routers with hashing to address load imbalance.

Recent studies have attempted to build attention-like multi-head routers.
Wu et al.~\cite{wuMultiHeadMixtureofExperts2024} propose using smaller FFNs to process sub-tokens directly, but this approach is unsuitable for upcycling scenarios and fails to highlight diverse token representation.
Another work \cite{huangMHMoEMultiHeadMixtureofExperts2024} tunes parameter settings for higher efficiency based on \cite{wuMultiHeadMixtureofExperts2024}.
Our method differs by initializing a mixture of collaborative routers from attention modules in a dense checkpoint, enhancing the model's ability to capture diverse patterns within the data and leading to improved performance and stability in MoE upcycling scenarios.

%% file: ecai/08-conclusion.tex
\section{Conclusion}
\label{sec: conclusion}

We introduce the first router specifically designed for upcycling in Mixture-of-Experts (MoE) models, employing a mixture of collaborative routers initialized from the attention module in the backbone dense model.
Our method enhances the routing mechanism's alignment and diversity by projecting tokens and experts into multiple low-dimensional representations and computing matching scores in an attention-like mechanism across these subspaces.
This framework pioneers router optimization in the upcycling scenario and extends upcycling from only upcycling the experts to upcycling the entire MoE structure.
Our future work will extend the method to other models and investigate the theoretical aspects of routing mechanisms in upcycling for deeper insights.